%% file: main.tex
\documentclass{article}
\usepackage{arxiv,times}

\usepackage[utf8]{inputenc} %
\usepackage[T1]{fontenc}    %
\usepackage{hyperref}       %
\usepackage{url}            %
\usepackage{booktabs}       %
\usepackage{amsfonts}       %
\usepackage{microtype}      %
\usepackage{xcolor}         %
\usepackage{wrapfig}
\usepackage{graphicx}

\usepackage{tabularx}

\usepackage{enumitem}

\usepackage{amsmath}
\usepackage{amssymb}
\usepackage{mathtools}
\usepackage{amsthm}
\usepackage{multirow}

\usepackage[capitalize,noabbrev]{cleveref}

\theoremstyle{plain}

\theoremstyle{definition}

\theoremstyle{remark}

\newcommand{\myparagraph}[1]{\noindent{\bf #1}}

\usepackage{xcolor}

\title{Scalable Delphi: Large Language Models\\for Structured Risk Estimation}

\author{%
  Tobias Lorenz \& Mario Fritz \\
  CISPA Helmholtz Center for Information Security, Germany \\
  \texttt{\{tobias.lorenz,fritz\}@cispa.de}
}

\begin{document}

\maketitle

\begin{abstract}
Quantitative risk assessment relies on structured expert elicitation to estimate unobservable properties. The Delphi method produces calibrated, auditable estimates but requires months of coordination and specialist time, placing rigorous risk assessment out of reach for most applications. We propose Scalable Delphi, adapting the classical protocol for LLMs with diverse expert personas, iterative refinement, and rationale sharing. Beyond lowering cost, this makes the assessment analyzable and dynamic. Rationales and revision histories record what each estimate rests on, information can be ablated to test which evidence matters, and the elicitation can be rerun with new evidence, changed assumptions, or adverse scenarios. Because target quantities are unobservable by construction, we design an evaluation framework based on necessary conditions any reliable estimator must satisfy: accuracy and calibration on verifiable proxies, and sensitivity to evidence. Agreement with expert panels and reasoning quality serve as corroboration. Across two domains (AI-augmented cybersecurity risk, ice-sheet contribution to sea-level rise), three benchmarks, and three reproduced expert studies, the estimates pass these tests: they improve systematically as evidence is added, agree with expert panels on most quantities, and correlate strongly with ground truth (Pearson $r = 0.91$--$0.98$).
\end{abstract}

\section{Introduction}
\label{sec:introduction}

What is the probability that a ransomware attack against a regional hospital succeeds? How likely is a semiconductor shortage to disrupt automotive production next quarter? How much will the ice sheets contribute to global sea level rise by 2100? How safe is a proposed small modular reactor design?
These questions span different domains, but they share a common structure: they are central to high-stakes decisions, yet they cannot be answered through direct measurement. We cannot run controlled experiments on real adversaries, simulate the full complexity of global supply chains, or observe the counterfactual where a reactor design fails. Instead, practitioners rely on structured expert judgment to produce calibrated estimates conditioned on available evidence.

The established methodology is the Delphi method~\citep{dalkey1963experimental}, developed at RAND in the 1950s and now standard in high-stakes domains. Studies proceed in rounds: experts receive briefing materials and independently provide probability estimates with written rationales; a facilitator anonymizes and circulates a summary noting agreement, disagreement, and outlier reasoning; experts then revise in light of this feedback until convergence. The result is a structured, auditable record of expert judgment that accounts for uncertainty and mitigates some forms of individual bias.

This rigor is expensive. A three-round Delphi study typically takes three to four months~\citep{gordon2009realtime} and requires coordinating multiple experts who may each charge several hundred dollars per hour. The NUREG-1150 risk assessment for five U.S. nuclear plants involved several hundred experts estimating hundreds of parameters over multiple years~\citep{nureg1990severe}. Such investments are justified when stakes are high enough, but they place quantitative risk assessment out of reach for most applications: small organizations cannot afford it, rapidly evolving domains cannot wait for it, and risk models with hundreds of interdependent parameters cannot be populated one expert panel at a time. And a completed study is a snapshot: when evidence changes or an assumption is challenged, there is rarely budget to reconvene the panel.

This paper investigates Large Language Models (LLMs) as panelists in structured expert elicitation. LLMs trained on academic literature, threat reports, and domain-specific documentation may have internalized enough knowledge and reasoning to provide useful probability estimates when given the same structured evidence as human experts. They have demonstrated calibration on forecasting tasks~\citep{kadavath2022language,halawi2024approaching} and can articulate reasoning that can be inspected and challenged. If these capabilities transfer to structured risk estimation, the economics change: elicitation becomes cheap enough to apply broadly, fast enough to keep pace with evolving threats, and reproducible enough to audit and version. Preliminary work has begun to probe this~\citep{nobrega2023aidelphi,mueller2024crafting,papakonstantinou2025privateaidelphi,barrett2025toward}, but LLMs can produce confident but poorly-grounded estimates and are not explicitly trained for probabilistic reasoning. Whether the resulting estimates are reliable is an empirical question.

Latent estimation differs from LLM forecasting on resolved future events~\citep{halawi2024approaching,schoenegger2024wisdom}. Forecasting asks for probabilities over outcomes that will eventually be observed, so a prediction can in principle be scored against the realized outcome. The probability that an attacker with specified capabilities succeeds against a defended target, or that sea level rises by a given amount under a specified emissions scenario, is instead a conditional rate over an assumption space: the same quantity can be re-estimated indefinitely as the assumptions are varied, any realized instance is consistent with a wide range of underlying rates, and direct ground-truth evaluation in the forecasting sense is not available.

Running the protocol with LLM panelists has multiple potential advantages: First, elicitation becomes cheap and fast enough to apply broadly and to keep pace with evolving threats. Second, it becomes analyzable. Every estimate comes with a rationale, a revision history, and a reproducible prompt, so an analyst can ask what an estimate depends on by ablating evidence, clarifying the question, varying the base model, or branching from the same context, rather than interpreting a single completed study. Third, it becomes dynamic. When evidence arrives, an assumption is disputed, or a worst case needs exploring, the relevant quantities can simply be elicited again, and a completed study becomes a reusable procedure rather than a snapshot.

Since latent quantities lack ground truth, validating estimates requires different machinery. We test whether LLM-based estimates exhibit properties any reliable estimator should satisfy: sensitivity to relevant evidence, accurate prediction of quantities we can verify, alignment with human expert judgment, and reasoning that engages the substance of the problem.

We instantiate this in two domains. The primary testbed is AI-augmented cybersecurity risk: threat models are well-specified and decomposable, three published benchmarks (BountyBench, Cybench, CyberGym) provide observable ground truth, and two recent human expert elicitation studies offer independent baselines~\citep{murray2025mapping,barrett2025toward}. We additionally test domain transfer by replicating the ice-sheet expert elicitation of~\citet{bamber2019sealevel}.

On the cybersecurity benchmarks, panel estimates correlate strongly with ground truth (Pearson $r = 0.91$--$0.98$) and degrade to noise when agent and benchmark identities are withheld. Across three reproduced elicitation studies~\citep{murray2025mapping,barrett2025toward,bamber2019sealevel}, estimates agree with expert panels on most quantities. The largest divergences occur where every panel is least certain: long-horizon extrapolations and open-ended count quantities. 
Inspecting rationales and reasoning traces allowed us to trace a systematic gap in the ransomware study to an ambiguity in how the target quantity was conditioned and allowed us to rerun the elicitation to produce consistent estimates. Rerunning the sea-level study with observations published after it likewise updates a completed assessment with evidence the original panel never saw, refining the estimates.

\section{Scalable Delphi Method}
\label{sec:method}
Structured risk models decompose complex risks into networks of conditional quantities, often formalized as fault trees, Bayesian networks, or influence diagrams~\citep{bedford2001probabilistic}. Each node requires an estimate, conditioned on explicit assumptions and consistent with the model's other nodes. Some quantities can be measured directly; many others cannot. For these latent quantities, structured expert elicitation is the standard approach~\citep{cooke1991experts}. We focus on the core subtask: producing calibrated estimates for individual quantities, which can then populate such models.

Delphi elicitation, as described in \cref{sec:introduction}, is the standard way to obtain such estimates but is costly, slow, and non-repeatable. We propose Scalable Delphi: given structured evidence and assumptions, a panel of LLM agents produces estimates through the same deliberation process.

\subsection{The Estimation Task}

Let $Q$ denote a quantity to be estimated in a risk model, and let $E$ denote the relevant evidence. The elicitation task is to produce an estimate $\hat{q}$ of $Q$ given $E$, together with a measure of uncertainty.

Mirroring traditional Delphi, we build a panel of $k$ expert agents. Each agent $j$ is instantiated with a distinct persona $\pi_j$ representing a particular perspective or expertise. These agents produce independent estimates for each round $r$, conditioned on $Q$, $E$, and the anonymized and aggregated feedback of the previous round $F^{(r-1)}$: $\hat{q}^{(r)}_j = \mathcal{M}(Q, E, \pi_j, F^{(r-1)})$.
$\mathcal{M}$ is the language model, and $F^{(0)} = \emptyset$ for the first round. The final estimate is aggregated across the panel after the last round.

\subsection{Elicitation Protocol}

We instantiate the full Delphi protocol with LLM agents in all roles: expert panelists \emph{and} mediator. This enables fully automated elicitation with no human involvement beyond problem specification.

\myparagraph{Expert Panel.}
We instantiate a panel of $k$ experts with personas $\Pi = \{\pi_1, \ldots, \pi_k\}$ chosen to reflect diverse perspectives on the task. Personas are the protocol's means of introducing the diversity that human panels draw from heterogeneous expertise~\citep{rowe1999delphi}. \Cref{sec:alignment} measures their effect alongside that of the base model.

\myparagraph{Round Structure.}
In round 1, each expert receives evidence $E$ and independently produces an estimate $\hat{q}^{(1)}_j$ along with a written rationale. A mediator then synthesizes the responses into feedback $F^{(1)}$: summary statistics of the estimates, key arguments for higher and lower values, and areas of agreement or disagreement, without attributing views to specific experts. In subsequent rounds, each expert receives this feedback and submits a revised estimate $\hat{q}^{(r)}_j$. This multi-round structure balances independent judgment with structured deliberation.

\myparagraph{Aggregation.} The panel estimate is a summary statistic of round-$R$ panelist estimates; the appropriate choice depends on the downstream application. We use the mean for point-valued quantities, the median for skewed distributions (\cref{sec:alignment}), and Cooke's equal-weight classical model for the full 5/50/95 percentile triples elicited in the Bamber replication \citep{cooke1991experts}.

\myparagraph{Prompt Structure.}
Prompts are organized into system-level and user-level components. The system prompt establishes: (1) the Delphi process context, explaining the expert's role; (2) the expert persona; and (3) output format requirements, including point estimates, plausibility ranges, and rationales. The user prompt provides the specific elicitation task: a description of the quantity to be estimated, the evidence $E$, and any relevant context. This separation allows the same evidence to be presented to multiple experts with different personas, and allows the same expert to be queried across multiple quantities. Prompt details are in \cref{app:prompts}.

\myparagraph{Repeatability and control.} Because every input to the protocol is explicit, an elicitation can be repeated under controlled changes. The evidence block can be replaced or ablated (\cref{sec:sensitivity}, \cref{app:bamber-evidence}), the question restated (\cref{sec:alignment}), the base model or personas exchanged, and the interaction history preserved or reset between quantities, trading coherence across related estimates against independence from earlier answers.

\section{Evaluation Framework}
\label{sec:evaluation}

The central question, whether structured LLM elicitation produces reliable estimates of latent quantities, cannot be answered by direct comparison to ground truth. In supervised learning we evaluate against held-out labels; in forecasting, against realized outcomes. Neither applies here: the probability that a novel exploit is found or that an adversary escalates from reconnaissance to attack lacks ground truth we can observe.

This does not make evaluation impossible, only indirect. We adopt a falsificationist and evidence-gathering stance: rather than asking whether LLM estimates are correct, we ask whether they fail tests that any reliable estimator must pass, and supplement this with corroborating evidence.

\myparagraph{Necessary Conditions.}
\textbf{\emph{Calibration:}} where ground truth is observable on proxy tasks of similar nature, estimates track true values and predicted probabilities match observed frequencies.
\textbf{\emph{Sensitivity:}} adding decision-relevant evidence moves estimates in the appropriate direction; removing it degrades them toward uninformed priors. Failure on either is disqualifying.

\myparagraph{Corroborating Evidence.}
\textbf{\emph{Agreement with expert panels:}}
where independent expert estimates are available, agreement suggests estimates track genuine properties of the domain rather than artifacts of model or protocol; disagreement is not automatically disqualifying since experts are fallible.
\textbf{\emph{Reasoning quality:}} 
where rationales are elicited alongside estimates, we inspect them to verify that reasoning engages provided evidence rather than generic patterns, and, where panels disagree, to locate the source of the disagreement.

Together, these evaluations provide the evidential basis for using the resulting estimates in practice.

\section{Experiments}
\label{sec:experiments}

We instantiate our evaluation framework primarily in the domain of AI-augmented cybersecurity. This domain offers three properties essential for our evaluation: benchmarks that provide observable ground truth (LLM agent success rates on security tasks), structured information that can be systematically varied, and two post-cutoff human expert studies that provide independent baselines~\citep{murray2025mapping,barrett2025toward}. We additionally probe domain transfer by replicating a sea-level-rise elicitation~\citep{bamber2019sealevel}. We test calibration and evidence sensitivity across three benchmarks, compare estimates to human expert panels, and evaluate with two frontier and three open models.

\subsection{Experimental Setup}
\label{sec:exp-setup}

\myparagraph{Benchmarks.}
We evaluate on three cybersecurity benchmarks with published agent success rates (Appendix~\ref{app:data}).
\textbf{BountyBench}~\citep{zhang2025bountybench} reports success rates for 10 agents across three task types (detection, exploitation, patching) on 25 real-world systems with 40 bug bounties.
\textbf{Cybench}~\citep{zhang2025cybench} reports success rates for 20 agents on 40 professional-level Capture the Flag tasks.
\textbf{CyberGym}~\citep{wang2025cybergym} reports success rates for 23 agents tasked to generate proof-of-concept exploits across 1,507 vulnerabilities in 188 open-source projects.

\myparagraph{Models.}
We evaluate two frontier models, GPT-5.1 (knowledge cutoff September 2024) and Claude Opus 4.1 (January 2025), as well as three open-weight models: GLM-5, Kimi K2.5, and Llama-4 Maverick. BountyBench and CyberGym were released after both frontier cutoffs, while an early version of Cybench predates both. \cref{app:contamination} reports direct tests for memorization.

\myparagraph{Metrics.}
We report Pearson correlation $r$, mean absolute error (MAE), expected calibration error (ECE, defined in \cref{app:calibration}), and how often the true value falls within the panelists' plausibility ranges, averaged across the panel.

\myparagraph{Baselines.}
We compare against simple heuristics: for BountyBench, \emph{task mean} (average across agents for each task type) and \emph{agent mean} (average across tasks for each agent); for Cybench and CyberGym, \emph{global mean} (leave-one-out average of all agents).

\subsection{Calibration}
\label{sec:calibration}

\begin{figure*}[t]
\centering

\includegraphics[width=\textwidth]{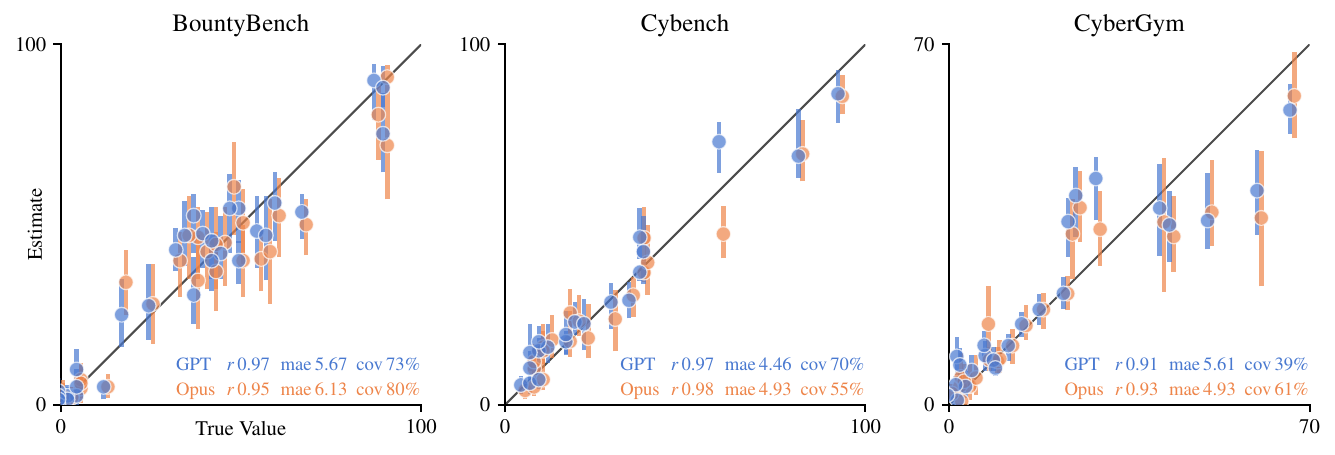}

\caption{Predicted vs.\ actual success rates across three cybersecurity benchmarks. Each point is the panel-mean estimate for a (model, task) cell; vertical bars span the panel-mean of agent plausibility ranges. Diagonal: perfect calibration. Inset: Pearson $r$ and mean absolute error per benchmark.}
\label{fig:calibration}
\end{figure*}

We test whether LLM estimates are calibrated with verifiable ground truth on proxy tasks.

\myparagraph{Task.} We use leave-one-out prediction tasks adapted to each benchmark's structure (full data in Appendix~\ref{app:data}). Cybench and CyberGym report aggregate success rates per agent (20 and 23 agents respectively); we hold out one agent and ask the LLM to estimate its overall success rate given the remaining agents and their descriptions. BountyBench reports success rates as a 10$\times$3 matrix of agents (rows) by task types (detection, exploitation, patching; columns); we hold out one matrix entry and ask the expert panel to estimate it given the remaining entries and the corresponding agent and task-type descriptions. For each prediction, we query a single round of five persona panelists and report the mean estimate.

\myparagraph{Results.}
\begin{table}[t]
\centering
\footnotesize
\setlength{\tabcolsep}{4pt}
\begin{tabular}{lrrrrrrrrr}
\toprule
& \multicolumn{3}{c}{BountyBench} & \multicolumn{3}{c}{Cybench} & \multicolumn{3}{c}{CyberGym} \\
\cmidrule(lr){2-4}\cmidrule(lr){5-7}\cmidrule(lr){8-10}
Model & $r$ & MAE & ECE & $r$ & MAE & ECE & $r$ & MAE & ECE \\
\midrule
GPT-5.1          & 0.97 & 5.7 & 1.6 & 0.97 &  4.5 & 4.1 & 0.91 & 5.6 & 2.6 \\
Opus-4.1         & 0.95 & 6.1 & 2.8 & 0.98 &  4.9 & 3.0 & 0.93 & 4.9 & 2.2 \\
GLM-5            & 0.96 & 6.3 & 2.2 & 0.97 &  4.2 & 4.0 & 0.94 & 5.0 & 2.1 \\
Kimi K2.5        & 0.96 & 5.9 & 2.5 & 0.98 &  4.0 & 2.3 & 0.92 & 5.8 & 3.4 \\
Llama-4 Maverick & 0.94 & 7.3 & 4.0 & 0.74 & 10.8 & 3.0 & 0.86 & 6.0 & 4.5\\
\bottomrule
\end{tabular}
\caption{Accuracy and calibration under full information. MAE and ECE in percentage points; ECE is task-weighted with 10 equal-width bins.}
\label{tab:accuracy}
\end{table}
Figure~\ref{fig:calibration} shows predicted versus actual success rates for the two frontier models, and Table~\ref{tab:accuracy} reports aggregates for all five models. Both frontier models achieve strong correlations with ground truth on all three benchmarks: on BountyBench, GPT-5.1 reaches $r = 0.97$ and Opus-4.1 $r = 0.95$; on Cybench, $r = 0.97$ and $0.98$; on CyberGym, $r = 0.91$ and $0.93$. Their mean absolute errors lie between 4.5 and 6.1 percentage points. The estimates are also well calibrated, with expected calibration errors between 1.6 and 4.1\,pp. On CyberGym, the largest deviation occurs in the highest bin, where the held-out agent outperforms every agent the panel was shown and the estimate must extrapolate beyond the given range.

The open-weight models show that these results do not strictly depend on frontier capability. GLM-5 and Kimi K2.5 perform on par with the frontier models on all three benchmarks. Llama-4 Maverick, the only model without extended reasoning, is close on BountyBench and CyberGym but noticeably worse on Cybench ($r = 0.74$, MAE 10.8\,pp).

All models substantially outperform simple heuristics. Predicting the mean of the remaining agents yields errors of 12.3, 20.0, and 16.7\,pp on the three benchmarks. On BountyBench, the task-type mean reaches a correlation of $r = 0.79$ by separating detection, exploitation, and patching, while the agent mean is uninformative ($r = -0.42$); the panels exceed both by also distinguishing between agents within each task type. Finally, the panelists' plausibility ranges contain the true value in 39 to 80\% of cases, depending on the benchmark. These ranges express the panelists' stated uncertainty and should not be read as calibrated confidence intervals.

\subsection{Evidence Sensitivity}
\label{sec:sensitivity}

\begin{figure*}[t]
\centering

\includegraphics[width=\textwidth]{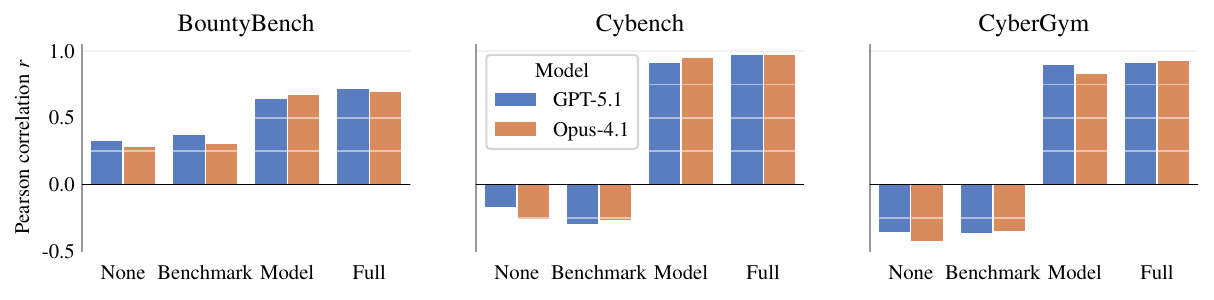}

\caption{Mean intra-task Pearson correlation across information conditions. Correlation increases as decision-relevant information is added, confirming estimates incorporate available evidence.}
\label{fig:sensitivity}
\end{figure*}

A reliable estimator must respond appropriately to available information. We test this by systematically varying the evidence provided to the model.

\myparagraph{Task.}
We repeat the leave-one-out prediction under four information conditions: (1) \emph{none}---anonymized agents and tasks, only numerical values; (2) \emph{benchmark}---benchmark description with anonymized agents; (3) \emph{model}---real agent names, anonymized benchmark; and (4) \emph{full}---both benchmark and agent information.

\myparagraph{Results.}
Figure~\ref{fig:sensitivity} shows mean intra-task Pearson correlation with ground truth across information conditions. Performance improves on all three benchmarks, with most of the gain coming from revealing agent identities; benchmark descriptions alone barely move correlation. Trajectories differ in shape: BountyBench has the most positive anonymized baseline ($r=0.30$) but the lowest full-information ceiling ($r=0.71$); the $10 \times 3$ matrix exposes patterns even under anonymization, but per-cell prediction across task types is harder than the aggregate-rate prediction on the other two benchmarks, where intra-task and overall correlation coincide. Cybench and CyberGym start negative ($r=-0.22$ and $-0.40$, respectively) and rise to $r=0.97$ and $r=0.92$ once both benchmark and agent information are supplied, a shift from noise to strong signal.

Rationales confirm this qualitatively: without evidence, experts interpolate from numerical patterns; with model identities they reason about relative capabilities (\cref{sec:qualitative}).

\subsection{Expert Alignment}
\label{sec:alignment}

The preceding experiments use benchmark success rates as ground truth. Here, we compare directly to human expert judgments. Suitable studies are scarce: comparison beyond headline numbers requires (1) published individual expert estimates, (2) ideally written rationales, and (3) a release date after the model training cutoff. Two cybersecurity studies meet all three \citep{murray2025mapping,barrett2025toward}. To probe domain transfer, we additionally replicate the landmark sea-level elicitation of \citet{bamber2019sealevel}; this predates the model cutoffs, but its per-expert data is ample and projection horizons extend to 2300, well beyond any plausibly memorized signal.

\begin{wrapfigure}{r}{0.4\linewidth}
    \vspace{-1.4em}
    \centering
    \includegraphics[width=\linewidth]{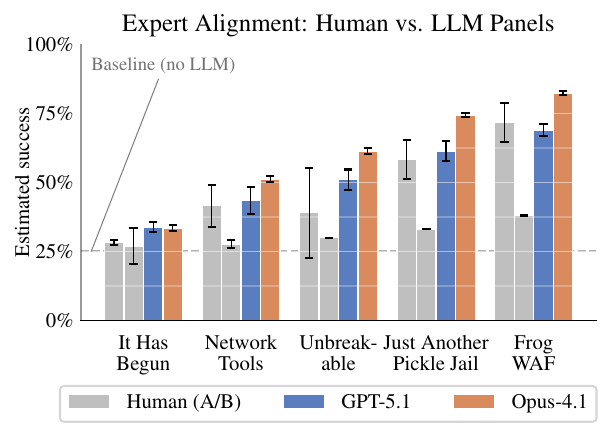}
    \caption{Expert alignment: LLM estimates compared to human expert panels from \citet{murray2025mapping}. Tasks ordered by difficulty based on human first-solve time. Bars show panel means; error bars indicate 95\% confidence intervals.}
    \label{fig:alignment1}
    \vspace{-1em}
\end{wrapfigure}

\myparagraph{Setup.}
We initiate our elicitation framework with a panel size of five, using diverse, domain-specific expert personas (\cref{app:personas}). We adjust our protocol to closely match the original studies. All three use dual-round elicitation. To match the human study conditions, each LLM expert maintains conversation history across tasks, allowing reference to previous estimates---just as human experts had context from earlier tasks in the sequence.

\myparagraph{\cite{murray2025mapping}}
asked cybersecurity experts to estimate the probability of successful malware development given access to an LLM of varying capability. Capability is operationalized through Cybench tasks: experts see a task description and estimate success probability, assuming that task represents the hardest challenge the LLM can solve (against a 25\% baseline with no LLM assistance). Five tasks of increasing difficulty (by human first-solve time) yield a mapping from benchmark performance to risk estimates. The study was released in March 2025, after both models' knowledge cutoffs.

\Cref{fig:alignment1} compares LLM panel estimates to the two human expert panels from \citet{murray2025mapping}. Both humans and LLMs show appropriate sensitivity to task difficulty: estimates increase monotonically from approximately 27--34\% for the easiest task to 38--82\% for the hardest.
The two human panels show substantial disagreement, with Panel~B providing systematically lower estimates than Panel~A (mean absolute difference (MAD) 16.6 percentage points (pp)). GPT-5.1 aligns closely with Human Panel~A (MAD 5.0~pp)---closer than the two human panels are to each other. Opus-4.1 estimates systematically higher risk (MAD 12.8~pp vs Human~A), but preserves the same relative ordering across tasks. The largest divergence occurs on ``Unbreakable,'' where LLMs estimate 51--61\% compared to 30--39\% for humans; notably, human experts in the original study flagged this task as having inflated first-solve time relative to its technical difficulty, which may explain both the high human variance and the models' higher estimates.

\begin{figure}[t]
    \centering
    \includegraphics[width=\textwidth]{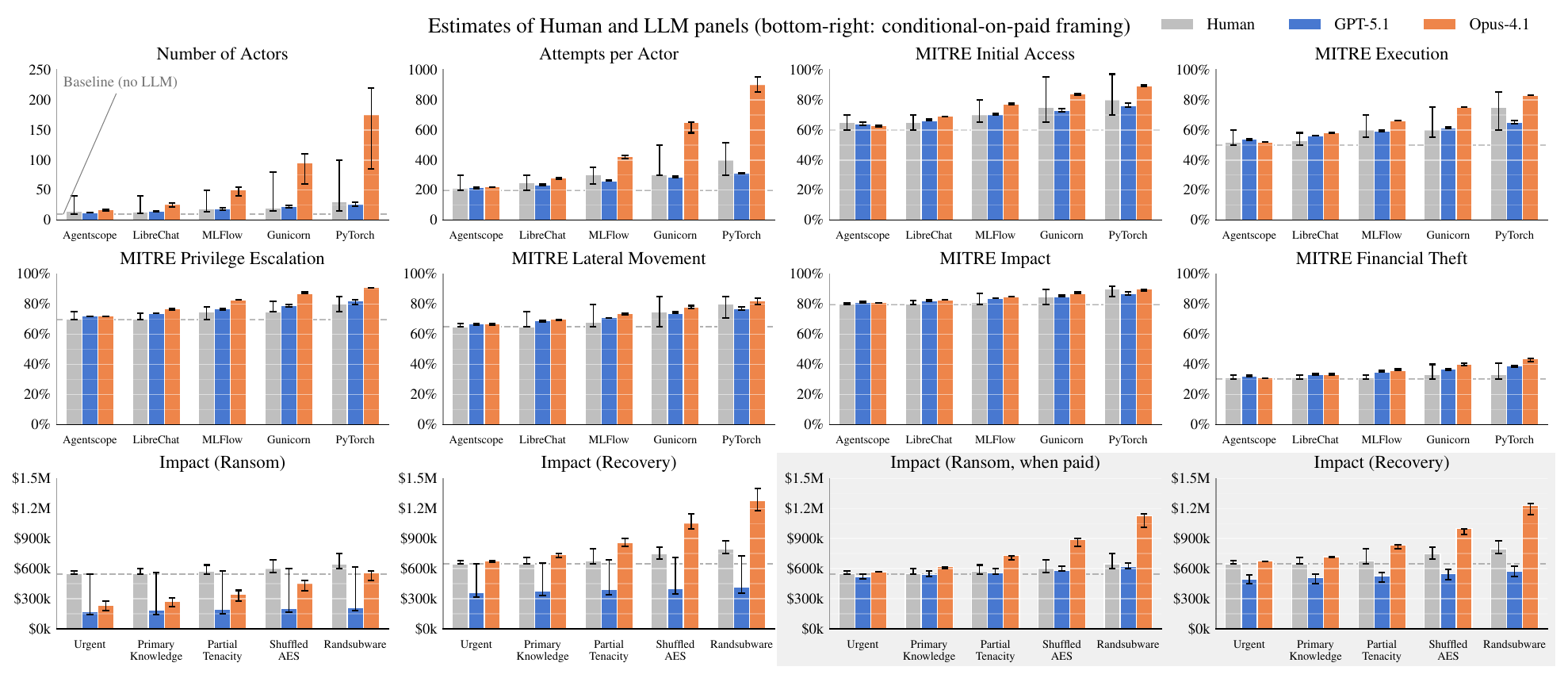}
\caption{Per-task estimates from a human panel \citep{barrett2025toward} and two agent panels (GPT-5.1, Opus-4.1) on the OC3 ransomware scenario. Bars are panel medians; error bars are bootstrap 95\% CIs of the median. The last two (gray) subplots repeat task 9 with disambiguated instructions.}
\label{fig:study2}
\vspace{-.7em}
\end{figure}

\myparagraph{\citet{barrett2025toward}}
build on this setting, broadening it to 10 quantities relevant to LLM uplift in ransomware. Experts are given pre-LLM baseline estimates and asked to predict the uplift when threat actors gain access to LLMs of varying capability, operationalized as the hardest-solve task across cybersecurity benchmarks. The study postdates both models' knowledge cutoffs.

\Cref{fig:study2} compares median estimates and bootstrap 95\% CIs across panels. LLM panels track human medians well on probability estimates, but diverge on counts (\emph{number of actors}, \emph{attempts per actor}): Opus-4.1 estimates greater uplift in the more capable scenarios, while GPT-5.1 stays closer to the human baseline throughout.

Both LLM panels anchor Impact (Ransom) below the pre-LLM baseline. Inspecting the rationales and reasoning traces reveals that the sheet lists a payment rate (30\%) and a conditional amount paid (\$550k). Four of five panelists in each panel start from their product, the expected value over all successful attacks, rather than from the conditional amount. Rerunning task 9 with the conditional quantity stated explicitly (gray, \cref{fig:study2}) removes this ambiguity and recovers the estimates. GPT-5.1 stays within 6\% of the human median at every tier, and Opus-4.1 starts close to those estimates, rising to $1.7\times$ at tier 5, the same upward extrapolation it shows on the count tasks. We analyze this case in detail in \cref{app:rat:09a}.

\begin{figure}[t]
    \centering
    \includegraphics[width=\textwidth]{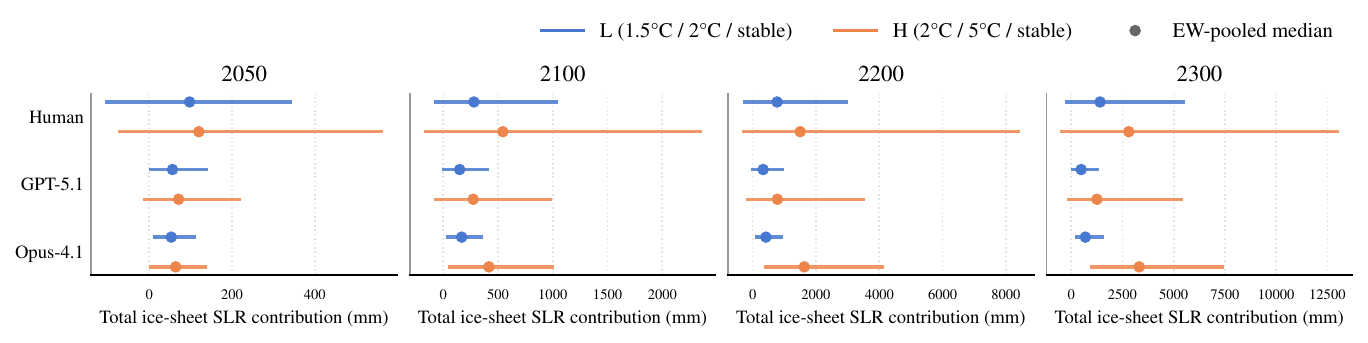}
\caption{\citet{bamber2019sealevel} SLR replication at matched $n{=}5$. Ice-sheet contribution to GMSL (mm, above 2000--2010 baseline) under low and high temperature trajectories. Filled circles and thin lines: per-panel medians and 5--95\% intervals, pooled across panelists via Cooke's equal-weight (EW) classical model \citep{cooke1991experts}. Human panel bootstrap-subsampled from $n{=}14$ to $n{=}5$ to match LLM panel size (\cref{fig:bamber-full-panel} shows the full-panel comparison).}
\label{fig:bamber-body}
\vspace{-1em}
\end{figure}

\myparagraph{\citet{bamber2019sealevel}}
elicited the contribution of the Greenland, West Antarctic, and East Antarctic ice sheets to global mean sea-level rise from 22 cryosphere experts using Cooke's classical structured expert judgment protocol \citep{cooke1991experts}. For each of 24 (ice-sheet, year, scenario) blocks, experts gave 5/50/95 percentile estimates of accumulation, runoff, and discharge anomalies above the 2000--2010 baseline, under low (L: 1.5\textdegree C $\to$ 2\textdegree C, stable) and high (H: 2\textdegree C $\to$ 5\textdegree C, stable) warming trajectories at four projection years (2050, 2100, 2200, 2300). Unlike the cyber studies, this paper predates both model cutoffs, so contamination cannot be excluded; we treat the comparison as a methodological complement, match panel size by bootstrap-sampling the $n{=}14$ valid human workbooks to $n{=}5$, and pool estimates via the Cooke equal-weight (EW) classical model.

\Cref{fig:bamber-body} shows the comparison across all three panels and four projection years. Both LLM panels track human medians at near-term horizons but diverge at long horizons. At 2050~H, GPT-5.1 and Opus-4.1 medians ($\sim\!65$~mm) are roughly half the human median (120~mm); Bamber's published performance-weighted PW01 median sits with the human EW pool. At 2300~H, GPT-5.1 remains near half while Opus-4.1 closes the gap, and the two LLM medians differ by $2.6\times$, comparable in magnitude to LLM-vs.-human disagreement. Despite these median-level divergences, all three panels share a structural similarity: right-skewed dispersion increases with year and warming.

\myparagraph{Evidence-aware re-elicitation.}
Because the elicitation can be rerun cheaply and without anchoring on earlier answers, a completed assessment can be updated when new evidence arrives. Rerunning the Bamber protocol with observations from after 2018 added to the briefing \citep{otosaka2023imbie} shifts the GPT-5.1 medians downward by 11 to 28\% across years and scenarios (Fig.~\ref{fig:bamber-evidence}, \cref{app:bamber-evidence}), demonstrating how assessments can be updated with new evidence.

\myparagraph{Trends.} Treating human estimates as a reference rather than ground truth, we find that across all three studies, LLM panels track human medians closely near known baselines (lower capability uplifts, shorter horizons) and diverge on farther extrapolations, where intra-panel variance also grows. Opus-4.1 trends more optimistic with larger extrapolations, GPT-5.1 more conservative; depending on the study, either can better match humans.

\myparagraph{Persona and model sensitivity.}
On Barrett, personas running on the same model differ by 2.4\,pp per task in the first round, while the same persona on different models differs by 4.8\,pp. Mediation reduces the first difference to 0.5\,pp but leaves the second at 4.4\,pp, so the choice of base model is the larger source of variation (App.~\ref{app:personas} reports a persona ablation on Murray).

\subsection{Rationales}
\label{sec:rationales}
Beyond numerical agreement, we examine \emph{why} each panel arrived at its estimates, using the human rationales available in \citet{barrett2025toward}. We use an LLM judge to label every rationale on a schema of 11 reasoning patterns. Rates, methodology, and a variance decomposition are in \cref{app:rationales}.

The three panels invoke the same constraint vocabulary at broadly comparable rates: technical and non-technical ceilings, baseline anchors, attacker-side framing, and hedging. The clearest difference is verbosity: human rationales average 234 characters, Opus 414, GPT 736. Consistent with that, GPT invokes every reasoning pattern more often than humans (mean $\Delta = +26.4$\,pp), while Opus tracks the human rate closely ($+8.8$\,pp) and falls \emph{below} humans on hedging, distributional reasoning, and tier-monotone justification. Two patterns diverge in the same direction for both LLM panels: role-marking and citation of the panel mean. Role-marking follows from persona prompting. Within-panel numeric convergence drops similarly for all three panels (Opus $0.49 \to 0.19$, humans $0.73 \to 0.55$, GPT $0.10 \to 0.08$), so humans update toward the panel mean too, they just do not write it down. The 11 patterns add only $\sim\!2\%$ of variance beyond panel and task fixed effects (\cref{app:rationales}).

\section{Discussion and Limitations}
\label{sec:discussion}

Scalable Delphi passes the necessary tests of our evaluation framework. Its estimates are accurate and calibrated on verifiable proxies across models, and they respond to the information available. The corroborating evidence points in the same direction: estimates align with expert panels directionally and often quantitatively, and rationales and reasoning traces show relevant arguments and perspectives.

\myparagraph{Scope.}
Our results support the method for its intended role: estimating specified quantities from a supplied evidence package to populate a risk model. They also show how to recognize estimates that deserve more scrutiny. Accuracy depends on the briefing, and disagreement concentrates where every panel is least certain, at long horizons and for open-ended counts. Both can be checked in a given run.

\myparagraph{Estimates versus uncertainty.}
Point estimates and aggregates are well calibrated, but the spread around them only partly tracks uncertainty. Within-panel spread indicates where disagreement is largest but not how large it is, and mediation narrows it further. This mirrors findings on LLM simulation of survey respondents, where average responses match while variance is compressed \citep{bisbee2024synthetic}. First-round spread and disagreement between base models are therefore better uncertainty signals than the final panel spread.

\myparagraph{Beyond a one-off panel.}
Two properties distinguish LLM elicitation from a traditional Delphi study. First, the record makes estimates open to analysis. In the ransom case, reasoning traces exposed ambiguous anchoring. Second, the elicitation can be rerun: clarifying the question resolved the ransom gap, and adding later observations updated the ice-sheet assessment, without reconvening a panel or anchoring on earlier answers. Whether panelists answer related quantities in a shared or a fresh context becomes a design choice. This leads to dynamic risk models that can be quickly and cheaply updated as evidence or assumptions change.

\myparagraph{Limitations.}
The target quantities have no observable ground truth, so our evaluation offers partial validation through proxy tasks, evidence sensitivity, and expert agreement. While indicative, none of these can deliver final proof; this limitation is inherent to the task and well established in the expert elicitation literature \citep{cooke1991experts}. Our evaluation spans two diverse domains. Extending it depends on published elicitations that release their briefing materials and per-expert estimates, which remain rare; as more such studies appear, the same framework applies to them directly.

\section{Related Work}
\label{sec:related}

\myparagraph{Structured Expert Elicitation.}
The Delphi method \citep{dalkey1963experimental,rowe1999delphi} established the core principles of structured elicitation: iterative rounds, anonymized feedback, and controlled interaction. Subsequent protocols formalized performance weighting \citep{cooke1991experts}, uncertainty quantification \citep{gosling2017shelf}, and structured aggregation \citep{hanea2017investigate}, and are standard in high-stakes domains \citep{nureg1990severe, bamber2013expert, mcandrew2021aggregating}.

\myparagraph{AI Risk Modeling.}
Quantitative risk modeling for AI systems is emerging \citep{touzet2025role, campos2025frontier}, with AI-enabled cyber offense among the most developed areas. Capability benchmarks (Cybench \citep{zhang2025cybench}, BountyBench \citep{zhang2025bountybench}, CyberGym \citep{wang2025cybergym}) measure LLM performance on security tasks, and expert elicitation studies translate these into risk estimates \citep{murray2025mapping, barrett2025toward}.

\myparagraph{LLM Forecasting.}
LLMs approach human crowd accuracy on resolved-event forecasting via retrieval \citep{halawi2024approaching} and ensembles \citep{schoenegger2024wisdom}, and calibration work characterizes when their probability estimates are reliable \citep{kadavath2022language, tian2023just}.

\myparagraph{LLM-based Deliberation.}
Multi-agent LLM systems improve reasoning through debate \citep{du2023improving, liang2024encouraging} and model negotiation \citep{abdelnabi2024cooperation}. Several studies apply LLMs in Delphi-style settings: qualitative future-of-work forecasting \citep{nobrega2023aidelphi}; scenario ratings showing moderate correlation with humans and positive bias \citep{mueller2024crafting}; AI-assisted nuclear reactor risk identification \citep{papakonstantinou2025privateaidelphi}; and cyber risk models using human and LLM-simulated panels \citep{barrett2025toward}. We share the last goal but focus on a rigorous evaluation framework rather than direct risk-model application.

\section{Conclusion}
\label{sec:conclusion}
We propose Scalable Delphi, which runs structured expert elicitation with LLM agents. Across three benchmarks and three reproduced expert studies, its estimates are accurate, calibrated, and responsive to evidence. Beyond lowering cost, it makes risk assessment analyzable and dynamic: estimates can be traced to their reasoning and updated as evidence changes.

\bibliography{references}
\bibliographystyle{arxiv}

\newpage
\appendix
\crefalias{section}{appendix}
\crefalias{subsection}{appendix}

\section*{Appendix}
\input{appendix/personas}
\input{appendix/calibration}
\input{appendix/qualitative-rationale-examples}
\input{appendix/bamber-full}
\input{appendix/appendix_bamber_evidence}
\input{appendix/appendix_rationale_analysis}
\input{appendix/appendix_contamination}
\input{appendix/prompts}
\input{appendix/appendix_cost}
\input{appendix/appendix_benchmarks}

\end{document}

%% file: appendix/personas.tex
\section{Personas}
\label{app:personas}

We run experiments with different sets of expert personas that fit the study domain. For the cyber uplift studies, we adapt the diverse security experts from \citet{barrett2025toward} (\cref{tab:personas}). For the \citet{bamber2019sealevel} replication, we design personas along similar axes, adjusted to the expertise of the topic at hand (\cref{tab:personas-slr}). For the ablation below, we use a uniform ``superforecaster'' personality.

\begin{table*}[h]
\small
\caption{Summary of expert personas for Delphi panels.}
\label{tab:personas}
\begin{tabularx}{\linewidth}{llXX}
\toprule
ID & Role & Background & Analytical Approach \\
\midrule
A & Defensive Security Specialist & 10 years SOC experience, APT detection & Defender's perspective, detection points \\
B & Malware Reverse Engineer & Anti-virus research lab & Bottom-up from technical implementation \\
C & AI/ML Security Researcher & PhD in CS, AI security & Systematic analysis of LLM assistance \\
D & Threat Intelligence Analyst & Former intelligence community & Observed attacker behavior patterns \\
E & Security Compliance Officer & CISSP, CISM certified & Framework-based, control effectiveness \\
\bottomrule
\end{tabularx}
\end{table*}

\begin{table*}[h]
\small
\caption{SLR Delphi panel personas (Bamber 2019 replication).}
\label{tab:personas-slr}
\begin{tabularx}{\linewidth}{lXXX}
\toprule
ID & Role & Background & Analytical Approach \\
\midrule
A & Greenland Outlet Glacier Dynamicist & 15 years tracking marine-terminating outlet glaciers with ICESat/ICESat-2 altimetry & Process-based ice-dynamics intuition anchored to observed retreat histories; emphasizes grounding-line retreat and dynamic discharge over SMB \\
B & Antarctic Ice Shelf Geophysicist & PhD in geophysics; field campaigns on Thwaites, Pine Island, Ross Ice Shelf & Reasons in terms of cavity geometry, basal melt rates, and grounding-line sensitivity; concerned with marine ice-sheet instability \\
C & Paleoclimate \& Sea-Level Reconstruction Specialist & Builds Holocene and LGM sea-level databases; collaborates with GIA modellers & Constrains estimates using proxy reconstructions and regional GIA adjustments; anchors variability to past rates \\
D & Ice-Sheet Numerical Modeller & Computational glaciologist running ISMIP6-class ensembles (PISM, CISM, ISSM) & Cross-checks observations against model-plausible ranges; widens intervals for structural uncertainty \\
E & Remote Sensing / Mass Balance Observations Lead & Leads a mass-balance observations group; author on IMBIE 2018/2019 reports & Direct observational anchoring (GRACE/GRACE-FO, CryoSat-2, IMBIE); cautious extrapolation beyond observed record \\
\bottomrule
\end{tabularx}
\end{table*}

\subsection{Persona Ablation}
\label{app:persona-ablation}

We compare the two persona configurations on \citet{murray2025mapping} (\cref{sec:alignment}): the original persona set in \cref{tab:personas}, and five uniform superforecaster personas, emphasizing calibrated probabilistic reasoning over domain-specific expertise. Aggregate estimates differ by 3.0~pp MAD for GPT-5.1 and 1.9~pp MAD for Opus-4.1; both small relative to human inter-panel disagreement (16.6~pp). The direction of difference varies by model.

%% file: appendix/calibration.tex
\section{Calibration}
\label{app:calibration}

\myparagraph{Definitions.}
For each held-out cell $i$, the panel predicts a success rate $\hat{q}_i$ and the benchmark reports the observed rate $q_i$ over $n_i$ tasks ($n_i = 40$ on BountyBench and Cybench, $1{,}507$ on CyberGym). We group cells into 10 equal-width bins on $[0, 100]$ by $\hat{q}_i$ and compute the expected calibration error as
\[
\mathrm{ECE} = \sum_{b} \frac{n_b}{N} \left| \bar{\hat{q}}_b - \bar{q}_b \right|,
\]
where $n_b = \sum_{i \in b} n_i$ is the number of task evaluations underlying the cells in bin $b$, $N = \sum_i n_i$, and $\bar{\hat{q}}_b$ and $\bar{q}_b$ are the task-weighted mean predicted and observed rates in the bin; each cell is thus weighted by its number of tasks, and $\bar{q}_b$ is the pooled observed rate of the bin. To judge whether an ECE value exceeds what sampling noise alone would produce, we simulate a perfectly calibrated panel: for each cell we draw an observed rate from $\mathrm{Binomial}(n_i, \hat{q}_i)/n_i$, recompute the ECE with the same bins, and repeat $10{,}000$ times. We report the central 95\% interval of this null distribution and the fraction of draws with an ECE at least as large as the observed one. On BountyBench and Cybench, where a cell rests on 40 tasks, the null interval is wide (upper bound 3.7--4.8\,pp) and every model's ECE falls inside it (GPT-5.1 on Cybench is the closest to the edge, $p = 0.10$). On CyberGym, where a cell rests on 1{,}507 tasks, the null interval is narrow (upper bound 0.6\,pp) and every model's ECE (2.1--4.5\,pp) lies above it: a small but detectable residual miscalibration.

\myparagraph{Brier scores.}
Table~\ref{tab:brier} reports the Brier score of each model, treating each benchmark task as a binary outcome with predicted probability $\hat{q}_i$. For a cell with observed rate $q_i$, the mean Brier score over its tasks equals $q_i(1-q_i) + (\hat{q}_i - q_i)^2$. The Murphy decomposition $\mathrm{Brier} = \mathrm{REL} - \mathrm{RES} + \mathrm{UNC}$ of the pooled score of the frontier models (0.129 for GPT-5.1, 0.128 for Opus-4.1, over all 73 cells and 36{,}661 task evaluations, 95\% of them on CyberGym) gives a reliability term of 0.006 against a resolution term of 0.040 and an uncertainty term of 0.163. This means the estimates gain considerably more from distinguishing between agents than they lose to miscalibration.

\begin{table}[h]
\centering
\footnotesize
\begin{tabular}{lrrr}
\toprule
Model & BountyBench & Cybench & CyberGym \\
\midrule
GPT-5.1          & 0.1556 & 0.1463 & 0.1276 \\
Opus-4.1         & 0.1576 & 0.1465 & 0.1270 \\
GLM-5            & 0.1571 & 0.1472 & 0.1259 \\
Kimi K2.5        & 0.1564 & 0.1457 & 0.1274 \\
Llama-4 Maverick & 0.1591 & 0.1703 & 0.1310 \\
\bottomrule
\end{tabular}
\caption{Brier scores under full information. Lower is better.}
\label{tab:brier}
\end{table}

%% file: appendix/qualitative-rationale-examples.tex
\section{Qualitative Rationale Examples}
\label{sec:qualitative}

To illustrate how evidence shapes reasoning, we compare rationales from the same expert on the leave-one-out calibration task of \cref{sec:calibration}, under two information conditions: anonymized numerical values only (\emph{none}), and full agent descriptions (\emph{full}).

\myparagraph{Without evidence.} Given only anonymized values, the expert resorts to statistical interpolation: ``The values form a rough performance ladder in increments of 2.5--5 points [\ldots] placing Model J at 32.5\% fits this progression.''

\myparagraph{With evidence.} Given model identities and descriptions, reasoning becomes substantive: ``GPT-4o is weaker than Claude 4.5 Sonnet (60\%) and likely below Claude 4.1/4 Opus (38\%), but clearly stronger than o3-mini (22.5\%), so an intermediate value around 30\% best fits the progression.''

Both estimates are reasonable, but evidence enables domain-informed comparison rather than blind extrapolation.

%% file: appendix/bamber-full.tex
\section{Bamber 2019: Full-Panel Comparison}
\label{app:bamber-full}

\begin{figure}[h]
    \centering
    \includegraphics[width=\textwidth]{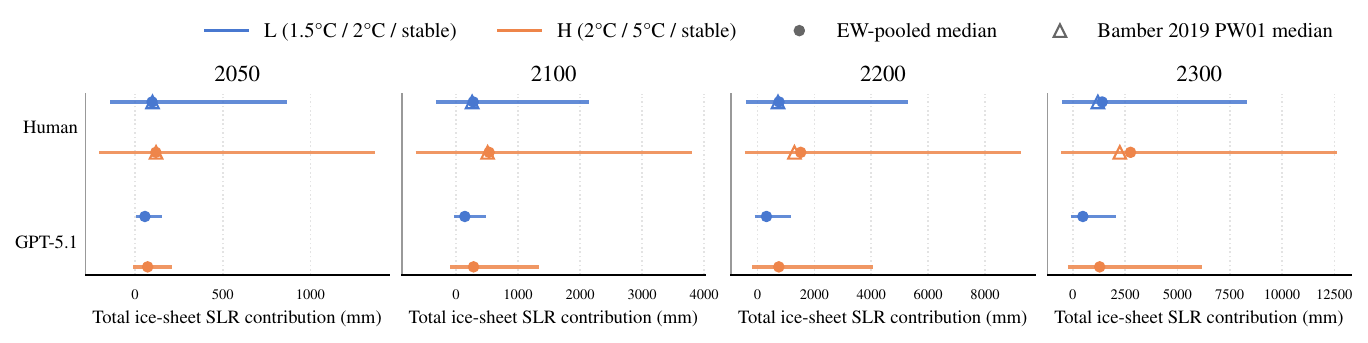}
    \caption{Bamber 2019 SLR replication, full human panel ($n{=}14$) vs. full GPT-5.1 panels ($n{=}14$). Otherwise identical to Figure~\ref{fig:bamber-body}.}
    \label{fig:bamber-full-panel}
\end{figure}

Figure~\ref{fig:bamber-body} in the main text matches LLM panel size by bootstrap-subsampling the $n{=}14$ valid human workbooks down to $n{=}5$. Figure~\ref{fig:bamber-full-panel} shows the same comparison at the full $n{=}14$ panel size for GPT-5.1; pooled medians and 5--95\% intervals are nearly indistinguishable from the matched-$n{=}5$ result, indicating the subsampling does not materially change the comparison.

%% file: appendix/appendix_bamber_evidence.tex
\section{Evidence-Aware Re-Elicitation}
\label{app:bamber-evidence}

We re-run the Bamber elicitation with post-2018 observational evidence prepended to the briefing, demonstrating that LLM Delphi panels can integrate observations arriving after the original study closed. For each ice sheet, the evidence is the cumulative SLR anomaly above the 2000--2010 baseline trend over the post-2018 window: Greenland from Mankoff's input-output method (2018--2023, $+0.1 \pm 0.6$~mm) and West and East Antarctic from \citet{otosaka2023imbie} (IMBIE 2018--2020, $+0.1 \pm 0.3$ and $+0.2 \pm 0.6$~mm). Antarctic windows are shorter because the post-2020 IMBIE per-sheet decomposition is not yet released; we trim Greenland to match. The evidence is added to the system prompt as a labeled block.

\begin{figure}[t]
\centering
\includegraphics[width=\textwidth]{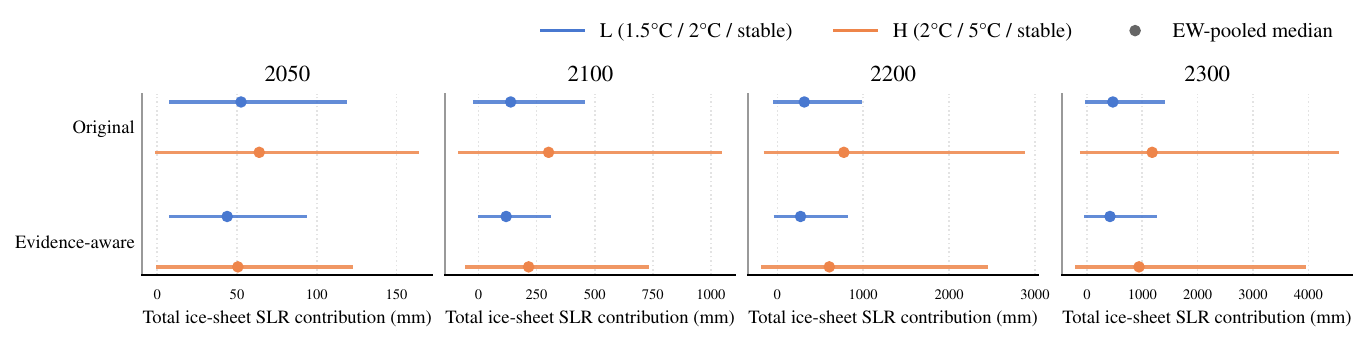}
\caption{\textbf{Effect of post-2018 observational evidence on the GPT-5.1 panel.} Total ice-sheet contribution to GMSL (mm above the 2000--2010 baseline trend) under low and high temperature trajectories, at four projection years. Per-panel 5/50/95 quantile triples pooled via Cooke's EW classical model \citep{cooke1991experts}; thin lines show the pooled 5--95 interval, dots the pooled median. ``Original'': no-evidence baseline. ``Evidence-aware'': a fresh Delphi run with post-2018 observed anomalies prepended. $n{=}5$ archetype personas.}
\label{fig:bamber-evidence}
\end{figure}

Figure~\ref{fig:bamber-evidence} compares GPT-5.1 with and without the evidence (otherwise identical: $n{=}5$ archetype personas, full Delphi protocol, Cooke EW pooling). Evidence-aware medians shift downward by $\sim$20\% across all years and scenarios (e.g., 2300~H: $1178 \to 941$~mm); the pooled 95th percentile compresses by 13\% at 2300~H. We make no normative claim about the magnitude of this shift, which depends on an implicit trajectory shape between 2005 and the projection year; the contribution is methodological---LLM Delphi panels can integrate post-elicitation observations as a structured pipeline step.

%% file: appendix/appendix_rationale_analysis.tex
\section{Cross-Panel Rationale Analysis}
\label{app:rationales}

\begin{table}
\centering
\footnotesize
\begin{tabular}{lccc}
\toprule
Reasoning & Human & GPT-5.1 $\Delta$ & Opus-4.1 $\Delta$ \\
\midrule
Invokes supplied pre-LLM baseline           & 15.4\% & \textcolor{green!60!black}{+12.2} & \textcolor{green!60!black}{+9.0}  \\
References values from previous steps       &  4.1\% & \textcolor{green!60!black}{+4.3}  & \textcolor{red!70!black}{-3.3}    \\
Names a technical ceiling                   & 25.3\% & \textcolor{green!60!black}{+55.1} & \textcolor{green!60!black}{+37.1} \\
Names a non-technical ceiling               & 13.7\% & \textcolor{green!60!black}{+28.1} & \textcolor{green!60!black}{+20.3} \\
Judges task relevance of benchmark          & 16.8\% & \textcolor{green!60!black}{+30.0} & \textcolor{red!70!black}{-0.2}    \\
Reasons about shape of distribution         &  1.9\% & \textcolor{green!60!black}{+11.9} & \textcolor{red!70!black}{-1.5}    \\
Uses attacker-side assumptions/reasoning    & 32.1\% & \textcolor{green!60!black}{+41.1} & \textcolor{green!60!black}{+18.5} \\
Defends change vs prior capability tiers    & 10.1\% & \textcolor{green!60!black}{+13.7} & \textcolor{red!70!black}{-7.9}    \\
Invokes expert's role or background         &  0.0\% & \textcolor{green!60!black}{+19.4} & \textcolor{green!60!black}{+10.0} \\
Cites panel mean or feedback                &  0.6\% & \textcolor{green!60!black}{+45.4} & \textcolor{green!60!black}{+49.4} \\
Substantive uncertainty hedging             & 65.8\% & \textcolor{green!60!black}{+29.0} & \textcolor{red!70!black}{-35.2}  \\
\midrule
\textbf{Mean}                               & \textbf{16.9\%} & \textcolor{green!60!black}{\textbf{+26.4}} & \textcolor{green!60!black}{\textbf{+8.8}} \\
\bottomrule
\end{tabular}
\vspace{.5em}
\caption{Per-panel rates pooled across round 1 and round 2. Human is the baseline; GPT-5.1 and Opus-4.1 columns show percentage-point deltas vs.\ Human.}
\vspace{-2em}
\label{tab:rat-headline-rates}
\end{table}

This appendix documents the labelling methodology behind \cref{tab:rat-headline-rates} and examines two cells where the panels disagreed sharply on the magnitude estimate: one where rationales locate the mechanism (Ransom impact, \cref{app:rat:09a}) and one where they do not (count tasks, \cref{app:rat:counts}).

\subsection{Labelling methodology}
\label{app:rat:method}

The corpus is 1{,}862 rationales pooled across human and LLM panels. LLM rationales come from the structured output of each agent; human rationales from the \texttt{Rationale} column of each expert's workshop sheet across both rounds. Tasks 09a (Impact--Ransom) and 09b (Impact--Recovery) share a single rationale per (agent, tier, round) by original study design. The human cell count falls below $9 \times 9 \times 5 \times 2 = 810$ singles plus 90 (duplicated 09a/09b) because 13\% of cells were left blank; we drop empty-rationale cells before classification.

Each rationale is classified by \texttt{gpt-5.5-2026-04-23} with \texttt{reasoning\_effort=medium} and structured output. The judge is in the GPT family but four minor versions ahead of the panel model (\texttt{gpt-5.1}); same-family bias on descriptive labels is small in our setting but not zero. Total judge cost: \$11.81 for 1{,}862 classifications.

The full schema contains 12 reasoning patterns. \cref{tab:rat-headline-rates} reports the 11 valid across tasks; we exclude consistency-math from the pooled summary because it fires exclusively on 09a/09b (\cref{app:rat:09a}).

A linear regression of relative uplift $(b - a)/a$ on the 11 patterns with panel and task fixed effects ($n = 1{,}862$) finds them adding 1.8\% of variance beyond panel and task identity ($R^2 = 0.271 \to 0.289$). The patterns describe \emph{what panels write}, not \emph{where the number lands}; the case studies illustrate both ends of this.

\subsection{Ransom impact (09a): anchoring on the expected-value line}
\label{app:rat:09a}

The workshop sheet for task 09 lists a payment rate (30\%) and a conditional amount paid (\$550k), while the risk-model question asks for the mean ransom across all successful attacks. Two anchors are therefore available: \$550k, conditional on payment, and $0.30 \times \$550\text{k} \approx \$165\text{k}$, the expected value over all successful attacks. Table~\ref{tab:ransom-r1} lists every panelist's round-1 best guess under the original and the disambiguated wording. We count a panelist as anchored on \$165k if its tier-1 estimate (the tier closest to the baseline) falls below the midpoint of the two anchors (\$357k). Under the original wording this holds for four of five panelists in each panel, and for none under the disambiguated wording. The human panel answered the conditional amount throughout.

\begin{table}[t]
\centering
\footnotesize
\setlength{\tabcolsep}{5pt}
\begin{tabular}{llrrrrr}
\toprule
Panel & Persona & Tier 1 & Tier 2 & Tier 3 & Tier 4 & Tier 5 \\
\midrule
\multirow{6}{*}{GPT-5.1, original}
 & A & 560 & 570 & 580 & 600 & 620 \\
 & B & 180 & 190 & 205 & 210 & 220 \\
 & C & 180 & 190 & 205 & 220 & 235 \\
 & D & 171 & 176 & 187 & 198 & 209 \\
 & E & 180 & 190 & 200 & 210 & 220 \\
 & \textit{median} & \textit{180} & \textit{190} & \textit{205} & \textit{210} & \textit{220} \\
\midrule
\multirow{6}{*}{Opus-4.1, original}
 & A & 575 & 605 & 660 & 740 & 825 \\
 & B & 175 & 220 & 330 & 475 & 625 \\
 & C & 180 & 230 & 320 & 450 & 575 \\
 & D & 170 & 185 & 220 & 300 & 400 \\
 & E & 170 & 195 & 240 & 310 & 390 \\
 & \textit{median} & \textit{175} & \textit{220} & \textit{320} & \textit{450} & \textit{575} \\
\midrule
\multirow{6}{*}{GPT-5.1, disambiguated}
 & A & 550 & 575 & 600 & 650 & 700 \\
 & B & 575 & 600 & 630 & 670 & 720 \\
 & C & 550 & 560 & 580 & 600 & 650 \\
 & D & 575 & 600 & 625 & 650 & 700 \\
 & E & 575 & 600 & 650 & 700 & 750 \\
 & \textit{median} & \textit{575} & \textit{600} & \textit{625} & \textit{650} & \textit{700} \\
\midrule
\multirow{6}{*}{Opus-4.1, disambiguated}
 & A & 565 & 590 & 650 & 750 & 900 \\
 & B & 575 & 650 & 800 & 1{,}000 & 1{,}400 \\
 & C & 580 & 660 & 850 & 1{,}100 & 1{,}400 \\
 & D & 575 & 600 & 670 & 780 & 950 \\
 & E & 560 & 590 & 680 & 820 & 1{,}000 \\
 & \textit{median} & \textit{575} & \textit{600} & \textit{680} & \textit{820} & \textit{1{,}000} \\
\midrule
Human, round 1 & \textit{median} & \textit{550} & \textit{550} & \textit{578} & \textit{632} & \textit{750} \\
\bottomrule
\end{tabular}
\caption{Round-1 best guesses (\$k) for Impact (Ransom) per persona, tiers 1--5 (Urgent, Primary Knowledge, Partial Tenacity, Shuffled AES, Randsubware), under the original and the disambiguated wording. Human round-1 medians (9 experts) for reference. The sheet lists a conditional amount paid (\$550k) and a payment rate (30\%). Under the original wording, four of five panelists in each panel start from their product (\$165k); under the disambiguated wording all start within 6\% of \$550k. Opus-4.1 panelists extrapolate more steeply with tier under both wordings.}
\label{tab:ransom-r1}
\end{table}

Agent reasoning traces show that under the original wording, every GPT-5.1 agent notices the sheet's inconsistency. Persona B reads the quantity as unconditional: ``The wording suggests the average economic damage from successful attacks also includes cases where zero ransom is paid. I need to apply Step 8's payout probability in calculating this.'' Its rationale carries the arithmetic: ``the baseline expected ransom impact ($\sim$\$165k) already folds in a $\sim$30\% payment rate and a mean paid amount around \$550k \ldots about \$220k.'' Persona A sees the same ambiguity and takes the other reading: ``I therefore keep the conditional mean ransom close to the baseline.'' Personas B--E anchor on \$165k, persona A on \$550k.

Opus-4.1 makes the same choice at the same rate, but only its reasoning traces show it. Persona C: ``The baseline values: \$550k ransom (with 30\% payment rate = \$165k expected).'' Persona D: ``Baseline: Ransom \$165k (30\% of \$550k).'' Persona E: ``\$165k baseline.'' Persona B starts at \$175k without naming a number; persona A alone takes ``the baseline of \$550k ransom (before 30\% payment rate).'' None of this reaches the written rationales, which speak only of ``slightly better data identification'' and ``modest increases,'' which is why the judge-coded analysis (App.~\ref{app:rationales}) finds no consistency-math for Opus. Each Opus panelist then applies percentage uplifts to its chosen baseline; the climb to \$625k at tier 5 is a large multiplier (``maximum achievable given constraints''), not a change of reading.

The final round pushed the correct reading out of the Opus panel. The mediator reported tier-1 ransom as ``min \$170,000, mean \$254,000, max \$575,000 \ldots one expert estimates notably higher.'' Opus persona A, the one panelist anchored on \$550k, moved from \$575k to \$275k: ``I was an outlier on the high side for ransom. Should moderate downward.'' GPT persona A held and named the cause: ``The panel's lower total-impact means likely reflect different framings (some mixing ransom size with payment probability), whereas I continue to treat this ransom figure as the conditional mean payout when victims do pay.'' This shows in \cref{fig:study2} Impact (Ransom), where GPT shows higher intra-panel spread compared to Opus.

Under the disambiguated wording both panels read the quantity as intended (Opus persona B: ``Ransom is CONDITIONAL on payment''; GPT persona A: ``the ransom mean is conditional on paying''), every tier-1 estimate is within 6\% of \$550k, and the remaining Opus-4.1 excess at tier 5 (\$1.4M for two personas) is an uplift belief (``40--80\% increase \ldots maximum extraction''), not an anchoring effect.

\subsection{Count tasks (01/02): rationales do not explain the divergence}
\label{app:rat:counts}

At tier 5 on the two count tasks (\cref{tab:rat-counts-r3}), the Opus/GPT ratio exceeds $2.9\times$, reaching $6.5\times$ on Number of Actors.

\begin{table}[h]
\centering
\caption{Round 2 panel medians on the count tasks at tier 5.}
\label{tab:rat-counts-r3}
\small
\begin{tabular}{@{}lcccc@{}}
\toprule
Task & Anchor & GPT-5.1 & Opus-4.1 & Human \\
\midrule
01 Number of Actors    &  10 &  27 & 175 &  30 \\
02 Attempts per Actor  & 200 & 310 & 900 & 400 \\
\bottomrule
\end{tabular}
\end{table}

Both LLM panels name a non-technical ceiling on all round 2 rationales for these cells (``RaaS capacity, law enforcement, market saturation''), at indistinguishable rates---the schema cannot account for the divergence. Round 1 panel means (\cref{tab:rat-r1-counts}) show the panels enter joint estimation already $\sim 6.5\times$ apart on Number of Actors and $3.1 \times$ on Attempts per Actor at tier 5.

\begin{table}[h]
\centering
\caption{Round 1 panel-mean estimates at tier 1 (easiest) and tier 5 (hardest) on count tasks.}
\label{tab:rat-r1-counts}
\small
\begin{tabular}{@{}lccccc@{}}
\toprule
Task & Tier & Anchor & GPT-5.1 & Opus-4.1 & Human \\
\midrule
01 Number of Actors    & 1 &  10 &  12 &  17 &  22 \\
01 Number of Actors    & 5 &  10 &  27 & 176 &  52 \\
02 Attempts per Actor  & 1 & 200 & 218 & 220 & 354 \\
02 Attempts per Actor  & 5 & 200 & 312 & 970 & 524 \\
\bottomrule
\end{tabular}
\end{table}

Both panels enumerate the same constraint set in round 1 but translate ``the LLM saturates the benchmark'' into different multipliers on the actor population. GPT reads saturation as ``near-senior automated exploit assistant'' (round 1 estimates imply a $\sim 2.7\times$ ceiling above baseline at tier 5); Opus reads it as ``democratisation of technical aspects'' ($\sim 18\times$ ceiling). The schema captures the appearance of constraint reasoning but not the magnitude of the ceiling, and on these tasks the ceiling is where the action is---a limitation of pattern-based rationale analysis.

%% file: appendix/appendix_contamination.tex
\section{Contamination Checks}
\label{app:contamination}

BountyBench (May 2025) and CyberGym (June 2025) were released after the knowledge cutoffs of both frontier models. An early version of Cybench (August 2024) predates both, so its initial results could be part of the training data. We run two direct tests on the frontier models.

\myparagraph{Recall without context.}
We ask each model to state the Cybench success rate of each agent without providing any other agents' results or descriptions. If the elicitation relied on memorized results, recall alone should be about as accurate. Instead, recall errors are 18.9\,pp for GPT-5.1 and 13.4\,pp for Opus-4.1, compared to 4.5 and 4.9\,pp for the elicitation, which is 4.2 and 2.7 times larger.

\myparagraph{Post-cutoff subset.}
We additionally restrict the Cybench evaluation to agents whose results were published after each model's cutoff: 11 agents for GPT-5.1 and 8 for Opus-4.1. The error changes from 4.5 to 5.3\,pp for GPT-5.1 and from 4.9 to 6.7\,pp for Opus-4.1, and Kendall's $\tau$ from 0.91 to 0.88 and from 0.89 to 0.91. All differences lie within bootstrap 95\% confidence intervals.

\myparagraph{Open-weight models.}
Llama-4 Maverick has a stated knowledge cutoff of August 2024, so BountyBench and CyberGym postdate it and Cybench coincides with it, the same situation as for the frontier models. GLM-5 and Kimi K2.5 were released after all three benchmarks, so the timing argument does not apply to them. We report the open-weight results as a test of whether performance depends on frontier capability, not as contamination-free evaluations.

%% file: appendix/prompts.tex
\section{Prompt Structures}
\label{app:prompts}
The expert system prompt establishes the Delphi/IDEA protocol context, describes the relevant scenario (e.g., malware development given LLM access), specifies the conditioning (e.g., $P(\text{success} \mid \text{access to capabilities})$ against a 25\% no-LLM baseline for the Murray replication), and defines the JSON output format (point estimate and rationale). The round 1 user prompt provides task-specific evidence (a Cybench task README for Murray, a workshop sheet for Barrett, ice-sheet/year/scenario blocks for Bamber) and asks the panelist for an initial estimate under the protocol's stated assumptions. The mediator receives anonymized round 1 estimates and rationales and produces a neutral summary of spread, agreement, and disagreement without recommending any value. The round 2 user prompt shows the mediator summary and asks panelists to revise or confirm with brief justification.

%% file: appendix/appendix_cost.tex
\section{Compute Resources}
\label{app:compute}

The frontier-model experiments use commercial LLM APIs (OpenAI GPT-5.1, Anthropic Claude
Opus~4.1). Tables~\ref{tab:compute-bench} and
\ref{tab:compute-studies} report API calls, token usage, cost, and wall-clock
time for the runs reported in the paper. Total reported spend is
approximately \$530 across 3{,}715 calls and \textasciitilde48M tokens. Including
preliminary and discarded runs, total project spend is approximately \$1{,}200.

\begin{table}[h]
\centering
\small
\caption{Calibration and information ablation experiments
(Sections~\ref{sec:calibration}--\ref{sec:sensitivity}; 3 benchmarks $\times$ 2 models).}
\label{tab:compute-bench}
\begin{tabular}{llrrrrrr}
\toprule
Run & Model & Calls & Input (M) & Output (M) & Reasoning & Cost (USD) & Wall (min) \\
\midrule
BountyBench & GPT-5.1   & 600 & 0.54 & 5.98 & 5.82 & \$118.76 & 34.0 \\
BountyBench & Opus-4.1  & 600 & 0.76 & 0.47 & --   & \$46.93  &  6.9 \\
Cybench     & GPT-5.1   & 400 & 0.36 & 2.76 & 2.66 & \$54.66  & 12.2 \\
Cybench     & Opus-4.1  & 400 & 0.53 & 0.28 & --   & \$29.13  &  3.3 \\
CyberGym    & GPT-5.1   & 460 & 0.50 & 3.14 & 3.02 & \$62.16  & 19.5 \\
CyberGym    & Opus-4.1  & 460 & 0.69 & 0.35 & --   & \$36.61  &  3.6 \\
\midrule
Subtotal    &           & 2{,}920 & 3.39 & 12.99 & 11.50 & \$348.26 & 79.6 \\
\bottomrule
\end{tabular}
\end{table}

\begin{table}[h]
\centering
\small
\caption{Human study replications (Section~\ref{sec:alignment}).}
\label{tab:compute-studies}
\begin{tabular}{llrrrrrr}
\toprule
Run & Model & Calls & Input (M) & Output (M) & Reasoning (M) & Cost (USD) & Wall (min) \\
\midrule
Murray      & GPT-5.1  &  55 & 0.23 & 0.07 & 0.07 & \$1.69  &  5.0 \\
Barrett & GPT-5.1  &  100 & 4.39 & 0.32 & 0.23 & \$10.94 & 10.7 \\
Barrett                       & Opus-4.1 &  100 & 4.56 & 0.10 & --   & \$75.59 & 10.8 \\
Bamber       & GPT-5.1  & 270 & 4.19 & 0.62 & 0.56 & \$16.99 & 23.5 \\
Bamber                        & Opus-4.1 & 270 & 4.29 & 0.13 & --   & \$74.01 & 13.2 \\
\midrule
Subtotal                      &          & 795 & 17.66 & 1.24 & 0.85 & \$179.22 & 63.3 \\
\bottomrule
\end{tabular}
\end{table}

%% file: appendix/appendix_benchmarks.tex
\section{Benchmark Data}
\label{app:data}

Tables~\ref{tab:bountybench-data}--\ref{tab:cybergym-data} show the ground-truth success rates and agent descriptions used in the leave-one-out calibration experiments (Section~\ref{sec:calibration}).

\begin{table*}[h]
\centering
\caption{BountyBench agent descriptions and success rates (\%). Success rates from \citet{zhang2025bountybench} (Apache-2.0); accessed on 2025-12-09 at \url{https://bountybench.github.io/}. Agent descriptions auto-generated from model cards and framework publications.}
\label{tab:bountybench-data}
\small
\begin{tabularx}{\textwidth}{@{}lXccc@{}}
\toprule
Agent & Description & Detect & Exploit & Patch \\
\midrule
Claude Code & Anthropic’s terminal-based agentic coding tool (Claude 3.7 Sonnet) with built-in code navigation and editing tools; optimized for software engineering and defensive tasks such as patching vulnerabilities. & 5.0 & 57.5 & 87.5 \\
\addlinespace[2pt]
OpenAI Codex CLI: o3-high & OpenAI’s Codex command-line coding agent using the o3-high reasoning model; can read, modify, and execute code with strong tool support, emphasizing reliable code understanding and patch generation. & 12.5 & 47.5 & 90.0 \\
\addlinespace[2pt]
OpenAI Codex CLI: o4-mini & Lightweight OpenAI Codex CLI variant using the o4-mini model; lower-cost, faster reasoning with the same coding-agent tool interface, optimized for efficient patching. & 5.0 & 32.5 & 90.0 \\
\addlinespace[2pt]
C-Agent: o3-high & Custom research agent based on the Cybench framework, running raw bash commands in Kali Linux with iterative memory and reasoning, powered by OpenAI’s o3-high model. & 0.0 & 37.5 & 35.0 \\
\addlinespace[2pt]
C-Agent: GPT-4.1 & Custom Cybench-style agent using OpenAI GPT-4.1, operating purely through terminal commands without specialized coding tools; balanced offensive and defensive behavior. & 0.0 & 55.0 & 50.0 \\
\addlinespace[2pt]
C-Agent: Gemini 2.5 & Custom Cybench-style agent powered by Google Gemini 2.5 Pro Preview, interacting via shell commands in Kali Linux for vulnerability exploitation and patching. & 2.5 & 40.0 & 45.0 \\
\addlinespace[2pt]
C-Agent: Claude 3.7 & Custom Cybench-style agent using Claude 3.7 Sonnet Thinking, emphasizing long-horizon reasoning for exploit development while retaining general patching capability. & 5.0 & 67.5 & 60.0 \\
\addlinespace[2pt]
C-Agent: Qwen2 35B A22B & Custom Cybench-style agent using Alibaba’s Qwen 35B A22B model, a large open-weight reasoning model executed via terminal-only interactions. & 0.0 & 17.5 & 25.0 \\
\addlinespace[2pt]
C-Agent: Llama 4 Maverick & Custom Cybench-style agent powered by Meta’s Llama 4 Maverick model, an open-weight multimodal-capable LLM used here in text-only terminal interactions. & 0.0 & 42.5 & 42.5 \\
\addlinespace[2pt]
C-Agent: DeepSeek-R1 & Custom Cybench-style agent using DeepSeek-R1, a reinforcement-learning–trained reasoning model designed for strong step-by-step problem solving in terminal environments. & 2.5 & 37.5 & 50.0 \\
\bottomrule
\end{tabularx}
\end{table*}

\begin{table*}[h]
\centering
\caption{Cybench agent descriptions and success rates (\%). Success rates from \citet{zhang2025cybench} (Apache-2.0); accessed on 2026-03-22 at \url{https://cybench.github.io/}. Agent descriptions auto-generated from model cards and framework publications.}
\label{tab:cybench-data}
\footnotesize
\begin{tabularx}{\textwidth}{@{}lXc@{}}
\toprule
Agent & Description & Success \\
\midrule
Claude 3 Opus & Large-scale Anthropic frontier model evaluated within the Cybench structured bash agent. Operates via iterative terminal commands in a Kali Linux environment to solve professional-level CTF challenges, measuring autonomous end-to-end exploitation without subtask guidance. & 10.0 \\
\addlinespace[2pt]
Claude 3.5 Sonnet & Anthropic’s high-performing general-purpose model used as the core reasoning engine in the Cybench agent. Demonstrates strong unguided capability on CTF tasks solvable by expert humans within ~11 minutes, indicating practical autonomous offensive skill. & 17.5 \\
\addlinespace[2pt]
Claude 3.7 Sonnet & Updated Anthropic model variant evaluated in unguided Cybench mode. Represents incremental improvements in reasoning and tool use over prior Claude 3.x models, though only unguided success is reported here. & 20.0 \\
\addlinespace[2pt]
Claude 4 Opus & Next-generation Anthropic flagship model evaluated on Cybench-style unguided CTF tasks. Higher success rate suggests substantially improved autonomous cybersecurity reasoning and execution compared to earlier Claude generations. & 38.0 \\
\addlinespace[2pt]
Claude 4 Sonnet & Anthropic mid-tier Claude 4 model evaluated in unguided Cybench tasks. Trades some raw capability for efficiency, but still demonstrates strong autonomous problem-solving on professional CTF challenges. & 35.0 \\
\addlinespace[2pt]
Claude 4.1 Opus & Refined Claude 4 Opus variant with improved reasoning robustness and continued gains in autonomous exploitation capability. & 38.0 \\
\addlinespace[2pt]
Claude 4.5 Sonnet & Advanced Claude 4.5 Sonnet model with strong unguided capability on professional CTF tasks solvable by expert human teams within short time horizons. & 60.0 \\
\addlinespace[2pt]
Gemini 1.5 Pro & Google DeepMind’s Gemini 1.5 Pro evaluated in Cybench’s unguided mode. Uses terminal-only interactions to solve CTF challenges; lower success reflects weaker autonomous exploitation relative to top Anthropic and OpenAI models. & 7.5 \\
\addlinespace[2pt]
GPT-4.5-preview & Preview OpenAI model evaluated in unguided Cybench setting. Represents an intermediate step toward stronger reasoning models, with limited reported metrics beyond overall unguided success. & 17.5 \\
\addlinespace[2pt]
GPT-4o & OpenAI’s multimodal GPT-4o used as the reasoning core of the Cybench agent. Demonstrates solid unguided success and is notable for solving tasks with higher human first-solve times when additional guidance is allowed. & 12.5 \\
\addlinespace[2pt]
Llama 3 70B Chat & Meta’s open-weight Llama 3 70B chat model evaluated in Cybench. Operates without specialized cybersecurity tooling; unguided results indicate limited autonomous exploitation capability. & 5.0 \\
\addlinespace[2pt]
Llama 3.1 405B Instruct & Large open-weight Meta model evaluated in Cybench unguided mode. Scale improves performance over smaller open models but remains behind leading closed models in autonomous CTF solving. & 7.5 \\
\addlinespace[2pt]
Mixtral 8x22B Instruct & Sparse Mixture-of-Experts model from Mistral evaluated in Cybench. Terminal-based agent shows modest unguided success, reflecting partial reasoning competence but weaker end-to-end execution. & 7.5 \\
\addlinespace[2pt]
OpenAI o1-mini & Small OpenAI reasoning-focused model evaluated in unguided Cybench mode. Designed for efficiency rather than maximal capability, yielding limited autonomous task completion. & 10.0 \\
\addlinespace[2pt]
OpenAI o1-preview & OpenAI reasoning-centric model evaluated in Cybench. While unguided success is modest, the model excels when subtask guidance is provided, indicating strong step-level reasoning. & 10.0 \\
\addlinespace[2pt]
OpenAI o3-mini & Compact OpenAI reasoning model evaluated in unguided Cybench tasks. Despite small size, shows improved autonomous problem solving over earlier compact models. & 22.5 \\
\addlinespace[2pt]
Grok 4 Fast & --- & 30.0 \\
\addlinespace[2pt]
Grok 4.1 Thinking & --- & 39.0 \\
\addlinespace[2pt]
Claude 4.5 Opus & --- & 82.0 \\
\addlinespace[2pt]
Claude Opus 4.6 & --- & 93.0 \\
\bottomrule
\end{tabularx}
\end{table*}

\begin{table*}[h]
\centering
\caption{CyberGym agent descriptions and success rates (\%). Success rates from \citet{wang2025cybergym} (Apache-2.0); accessed on 2026-03-22 at \url{https://www.cybergym.io/}. Agent descriptions auto-generated from model cards and framework publications.}
\label{tab:cybergym-data}
\scriptsize
\begin{tabularx}{\textwidth}{@{}lXc@{}}
\toprule
Agent & Description & Success \\
\midrule
Anthropic Agent: Claude-Opus-4.6 & Anthropic's strongest model (Feb 2026) with 1M-token context, agent teams, and a 14.5-hour task horizon, paired with Anthropic's own agent scaffolding; excels at multi-step planning, debugging, and extended autonomous coding workflows. & 66.6 \\
\addlinespace[2pt]
SageAgent: GPT-5 & OpenSage's self-programming agent framework---capable of generating agent topology, synthesizing tools, and managing hierarchical memory at runtime---combined with OpenAI's GPT-5 (Aug 2025), a frontier model with state-of-the-art coding and dynamic reasoning. & 60.2 \\
\addlinespace[2pt]
Anthropic Agent: Claude-Opus-4.5 & Anthropic's previous flagship (200K context, 64K output) with strong enterprise reasoning, programmatic tool calling, and tool-search capabilities, run through Anthropic's agent scaffolding for autonomous multi-step tasks. & 50.6 \\
\addlinespace[2pt]
Claude Code: GLM-5 & Zhipu AI's 744B-parameter MoE model (Feb 2026, $\sim$44B active, MIT-licensed, Ascend-trained) with native agent mode and strong coding abilities, driven by Anthropic's open-source Claude Code CLI for terminal-based file editing and code execution. & 43.2 \\
\addlinespace[2pt]
Kimi Agent: Kimi K2.5 & Moonshot AI's 1T-parameter MoE model (32B active, Jan 2026) with native multimodal understanding, 256K context, and Agent Swarm technology coordinating up to 100 specialized agents, paired with Kimi Code CLI for agentic task execution. & 41.3 \\
\addlinespace[2pt]
Anthropic Agent: Claude-Sonnet-4.5 & Anthropic's high-performance mid-tier model (200K context, 64K output) noted for strong coding, agent building, and computer use, with reduced sycophancy; run through Anthropic's agent scaffolding. & 28.9 \\
\addlinespace[2pt]
Anthropic Agent: Claude-Opus-4.1 & Anthropic's Opus 4.1 model with deep reasoning and strong tool-use capabilities for complex multi-step tasks, paired with Anthropic's agent scaffolding for autonomous execution. & 25.0 \\
\addlinespace[2pt]
Claude Code: GLM-4.7 & Zhipu AI's $\sim$400B-parameter model (Dec 2025) with 200K context, 128K output, and Interleaved Thinking for stable agentic orchestration; strong multilingual coding abilities driven by Claude Code CLI. & 23.5 \\
\addlinespace[2pt]
Anthropic Agent: Claude-Sonnet-4 & Anthropic's general-purpose Sonnet 4 model with solid reasoning and tool-use abilities, run through Anthropic's agent scaffolding for autonomous multi-step workflows. & 22.6 \\
\addlinespace[2pt]
OpenHands: Claude-Sonnet-4 & Anthropic's Sonnet 4 model paired with the OpenHands open-source software development agent framework, providing terminal, file editing, and browser tools for autonomous task execution. & 17.9 \\
\addlinespace[2pt]
Anthropic Agent: Claude-Sonnet-3.7 & Earlier Anthropic model optimized for coding and reasoning, run through Anthropic's agent scaffolding; capable but superseded by later Sonnet and Opus releases. & 14.5 \\
\addlinespace[2pt]
OpenHands: Claude-Sonnet-3.7 & Anthropic's Sonnet 3.7 model with strong coding and reasoning capabilities, paired with OpenHands' open-source agent framework for autonomous software development tasks. & 11.9 \\
\addlinespace[2pt]
OpenHands: GPT-4.1 & OpenAI's GPT-4.1 (Apr 2025) with up to 1M-token context and strong coding and instruction-following, paired with OpenHands' open-source agent framework. & 9.4 \\
\addlinespace[2pt]
Cybench: GPT-4.1 & Stanford CRFM's Cybench agent framework---designed for evaluating cybersecurity capabilities with subtask decomposition and professional-level CTF tasks---running GPT-4.1 as the underlying model. & 9.0 \\
\addlinespace[2pt]
Codex: GPT-4.1 & OpenAI's open-source Rust-based Codex CLI coding agent providing local terminal execution, file editing, and code generation, backed by GPT-4.1's strong coding and long-context capabilities. & 7.4 \\
\addlinespace[2pt]
ENiGMA: GPT-4.1 & ENiGMA (Enhanced Interactive Generative Model Agent), a cybersecurity-focused agent built on SWE-agent with novel interactive tools (debugger, server connections) and specialized agent-computer interfaces, powered by GPT-4.1. & 7.2 \\
\addlinespace[2pt]
OpenHands: Gemini-2.5-Flash & Google's Gemini 2.5 Flash model optimized for speed and efficiency rather than deep reasoning, paired with OpenHands' open-source agent framework for autonomous task execution. & 4.8 \\
\addlinespace[2pt]
OpenHands: DeepSeek-V3 & Large open-weight general-purpose model emphasizing efficient reasoning at scale, paired with OpenHands' open-source agent framework for autonomous software development tasks. & 3.6 \\
\addlinespace[2pt]
OpenHands: o4-mini & OpenAI's small, cost-efficient reasoning model with conservative agent behavior, paired with OpenHands' open-source agent framework for autonomous task execution. & 2.5 \\
\addlinespace[2pt]
OpenHands: R2E-Gym-32B & Open-weight 32B model (Qwen-based) trained on R2E-Gym trajectories with procedural environment generation and hybrid verifiers for software engineering tasks, run through OpenHands' agent framework. & 2.0 \\
\addlinespace[2pt]
OpenHands: Qwen3-235B-A22B & Very large open-weight Qwen3 MoE model ($\sim$235B total, $\sim$22B active) with broad general reasoning capacity, paired with OpenHands' open-source agent framework. & 1.9 \\
\addlinespace[2pt]
OpenHands: OpenHands-LM-32B & Open-weight 32B model fine-tuned from Qwen 2.5 Coder for the OpenHands framework using RL-based training on successful agent trajectories; optimized for software development with 128K context. & 1.7 \\
\addlinespace[2pt]
OpenHands: SWE-Gym-32B & Open-weight 32B Qwen 2.5 Coder model fine-tuned on agent-environment interaction trajectories from SWE-Gym's executable Python task environments, run through OpenHands' agent framework. & 0.1 \\
\bottomrule
\end{tabularx}
\end{table*}